\documentclass[letterpaper]{article}
\pdfoutput = 1
\usepackage{aaai}
\usepackage{times}
\usepackage{helvet}
\usepackage{courier}
\usepackage{graphicx}
\usepackage{amsmath,amssymb}

\usepackage{hyperref}

\begin{document}
%
\title{Keyphrase Generation for Scientific Articles using GANs}
\author{Avinash Swaminathan \textsuperscript{\rm 1}, Raj Kuwar Gupta\textsuperscript{\rm 1}, Haimin Zhang\textsuperscript{\rm 2}, \\ \bf \Large  Debanjan Mahata \textsuperscript{\rm 2},  Rakesh Gosangi\textsuperscript{\rm 2}, Rajiv Ratn Shah\textsuperscript{\rm 1} \\ 
\textsuperscript{\rm 1}MIDAS, IIIT-Delhi, India \\ \textsuperscript{\rm 2}Bloomberg \\ 
s.avinash.it.17@nsit.net.in, rajkuwargupta1996@gmail.com, hzhang449@bloomberg.net, \\ dmahata@bloomberg.net,  rgosangi@bloomberg.net, rajivratn@iiitd.ac.in 
}

\maketitle
\begin{abstract}
\begin{quote}
In this paper, we present a keyphrase generation approach using conditional Generative Adversarial Networks (GAN). In our GAN model, the generator outputs a sequence of keyphrases based on the title and abstract of a scientific article. The discriminator learns to distinguish between machine-generated and human-curated keyphrases. We evaluate this approach on standard benchmark datasets. Our model achieves state-of-the-art performance in generation of abstractive keyphrases and is also comparable to the best performing extractive techniques. We also demonstrate that our method generates more diverse keyphrases and make our implementation publicly available\footnote{Code is available at \href{https://github.com/avinsit123/keyphrase-gan}{https://github.com/avinsit123/keyphrase-gan}}.
\end{quote}
\end{abstract}

\section{Introduction}
\noindent Keyphrases are employed to capture the most salient topics of a long document and are indexed in databases for convenient retrieval. Researchers annotate their scientific publications with high quality keyphrases to ensure discoverability in large scientific repositories. Keyphrases could either be extractive (part of the document) or abstractive. Keyphrase generation is the process of predicting both extractive and abstractive keyphrases from a given document. This process is similar to abstractive summarization but instead of a summary the models generate keyphrases. 

Researchers have achieved considerable success in the field of abstractive summarization using conditional-GANs \cite{summaryIntro}. There has also been growing interest in deep learning models for keyphrase generation \cite{meng2017deep,Chan2019NeuralKG}. Inspired by these advances, we propose a new GAN architecture for keyphrase generation where the generator produces a sequence of keyphrases from a given document and the discriminator distinguishes between human-curated and machine-generated keyphrases.

\section{ Proposed Adversarial Model }
As with most GAN architectures, our model also consists of a generator (G) and discriminator (D), which are trained in an alternating fashion \cite{goodfellow2014generative}. 

\noindent \textbf{Generator} - 
Given a document $d = \{x_1, x_2, ..., x_n\}$, where $x_i$ is the $i\textsuperscript{th}$ token, the generator produces a sequence of keyphrases: $y = \{y_1, y_2, ..., y_m\}$, where each keyphrase $y_i$ is composed of tokens {${y_i^1, y_i^2, ..., y_i^{l_i}}$}. We employ catSeq model \cite{Yuan2018GeneratingDN} for the generation process, which uses an encoder-decoder framework: the encoder being a bidirectional Gated Recurrent Unit (bi-GRU) and the decoder a forward GRU. To incorporate the out-of-vocabulary words, we use a copying mechanism \cite{gu2016incorporating}. We also make use of attention mechanism to help the generator identify the relevant components of the source text. 

\noindent \textbf{Discriminator} - We propose a new hierarchical-attention model as the discriminator, which is trained to distinguish between human-curated and machine-generated keyphrases. The first layer of this model consists of $m+1$ bi-GRUs. The first bi-GRU encodes the input document $d$ as a sequence of vectors: $h = \{h_1, h_2, ..., h_n\}$. The other $m$ bi-GRUs, which have the same weight parameters, encode each keyphrase as a vector: $\{k_1, k_2, ..., k_m\}$. We then use an attention-based approach \cite{luong-etal-2015-effective} to build context vectors $c_j$ for each keyphrase, where $c_j$ is a weighted average over $h$. By concatenating $c_j$ and $k_j$, we get a contextualized representation $e_j = [c_j;k_j]$ of keyphrase $y_j$.
    

    

The second layer of the discriminator is another bi-GRU which consumes the document representation $h$ and the keyphrase representations $e$. The final state of this layer is passed through one fully connected layer ($W_f$) and sigmoid transformation to get the probability that a given keyphrase sequence is human-curated.
 
\begin{center}
    \tiny{    $s_t = 
    \begin{cases}
    GRU(h_t,s_{t-1}) ,\textbf{for  } 1\leq t\leq n \\
    GRU(e_{t-n}, s_{t-1}) , \textbf{  for  } n+1\leq t\leq n + m \\
     \end{cases}$ } \newline
    
    $R(y_{i}) = D(y_{i}) = \sigma(W_f s_{i+n}) $
\end{center}


\begin{figure}
\includegraphics[width=\linewidth,height=7cm]{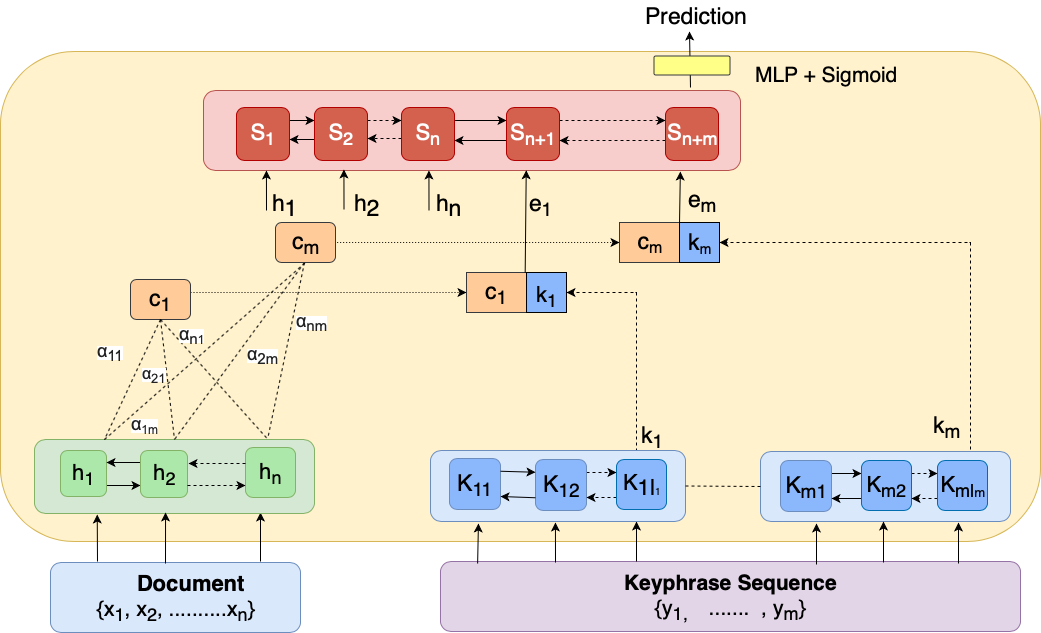}
\caption{Schematic of Proposed Discriminator(D) }
\label{fig:D_model}
\end{figure}

\noindent \textbf{GAN training} - For a given dataset (S), which contain the documents and corresponding keyphrases, we first pre-train the generator (G) using Maximum Likelihood Estimation. We then use this generator to produce machine-generated keyphrases for all documents in S. These generated keyphrases along with the curated keyphrases are used to train the first version of the discriminator (D).

We then employ policy gradient reinforcement learning to train the subsequent versions of G. We freeze the weight parameters of D and use it for reward calculation to train a new version of G. The reward for each keyphrase is obtained from the last $m$ states of the second bi-GRU layer in D (see Figure \ref{fig:D_model}). The gradient update is given as:
\begin{center}
\tiny{$\bigtriangledown_{}R_G = \sum_{i=1}^{m}[D(y_{i})-B] \bigtriangledown log  \prod_{j=1}^{l_{i}} G( y_{i}^{j}$ | $y_{i}^{1:j-1},  y_{1:i-1}, x)$}

\end{center}

\noindent where B is a baseline obtained by greedy decoding of keyphrase sequence. The resulting generator is then used to create new training samples for D. This process is continued till G converges. 

\section{Experiments and Results}
We trained the proposed GAN model on KP20k dataset \cite{meng2017deep} which consists of 567,830 samples for training, 20,000 each for testing and validation. Each sample consists of an abstract, title, and the corresponding keyphrases of a scientific article. We evaluated the model on four datasets: Inspec, NUS, KP20k, and Krapivin, which contain 600, 211, 20,000, and 800 test samples respectively. For training G, we used Adagrad optimizer with learning rate $\approx$ 0.0005. We compare our proposed approach against 2 baseline models - catSeq \cite{Yuan2018GeneratingDN}, RL-based catSeq Model \cite{Chan2019NeuralKG} in terms of F1 scores as explained in \cite{Yuan2018GeneratingDN}. The results, summarized in Table \ref{tab:tab1l}, are broken down in terms of performance on extractive and abstractive keyphrases.

For extractive keyphrases, our proposed model performs better than the pre-trained catSeq model on all datasets but is slightly worse than catSeq-RL except for on Krapivin where it obtains the best F1@M of 0.37. On the other hand, for abstractive keyphrases, our model performs better than the other two baselines on three of four datasets suggesting that GAN models are more effective in generation of keyphrases.


We also evaluated the models in terms of $\alpha$-nDCG@5 \cite{NDCG}. The results are summarized in Table \ref{tab:tab2}. Our model obtains the best performance on three out of the four datasets. The difference is most prevalent in KP20k, the largest of the four datasets, where our GAN model (at 0.85) is nearly 5\% better than both the other baseline models.

\section{Conclusion}
In this paper, we propose new GAN architecture for keyphrase generation. The proposed model obtains state-of-the-art performance in generating abstractive keyphrases. To our knowledge, this is the first work that applies GANs to keyphrase generation problem. 
\begin{table}[htbp]
    \centering
\resizebox{\columnwidth}{!}{%
\begin{tabular}{|l|l|l|l|l|l|}
\hline

Model & Score & Inspec & Krapivin & NUS & KP20k\\ \hline
Catseq(Ex) & F1@5 & 0.2350 & 0.2680 & 0.3330 & 0.2840  \\
           & F1@M & 0.2864 & 0.3610 & 0.3982 & 0.3661   \\ \hline
catSeq-RL(Ex.) & F1@5 & \textbf{0.2501} & \textbf{0.2870} & \textbf{0.3750} & \textbf{0.3100}   \\ 
& F1@M & \textbf{0.3000} & 0.3630 & \textbf{0.4330} & \textbf{0.3830}   \\ \hline
GAN(Ex.) & F1@5 & 0.2481 & 0.2862 & 0.3681 & 0.3002   \\ 
& F1@M & 0.2970 & \textbf{0.3700} & 0.4300 & 0.3810   \\ \hline
catSeq(Abs.) & F1@5 & 0.0045 & 0.0168 & 0.0126 & 0.0200   \\ 
& F1@M & 0.0085 & 0.0320 & 0.0170 & 0.0360   \\ \hline
catSeq-RL(Abs.) & F1@5 & 0.0090 & \textbf{0.0262} & 0.0190 & 0.0240    \\ 
& F1@M & 0.0017 & \textbf{0.0460} & 0.0310 & 0.0440    \\ \hline
GAN(Abs.) & F1@5 & \textbf{0.0100} & 0.0240 & \textbf{0.0193} & \textbf{0.0250}    \\ 
& F1@M & \textbf{0.0190} & 0.0440 & \textbf{0.0340} & \textbf{0.0450}   \\ \hline           
    
\end{tabular}
}
    \caption{\small{Extractive and Abstractive Keyphrase Metrics}}
    \label{tab:tab1l}
\end{table}

\begin{table}[htbp]
    \centering
    \scalebox{1}{
\begin{tabular}{|l|l|l|l|l|}
\hline

Model & Inspec & Krapivin & NUS & KP20k\\ \hline
Catseq & 0.87803 &  0.781 & 0.82118 & 0.804 \\ \hline
Catseq-RL & 0.8602 & \textbf{0.786} & 0.83 & 0.809\\ \hline
GAN & \textbf{0.891} & 0.771 & \textbf{0.853} & \textbf{0.85}\\ \hline
\end{tabular}}
\caption{$\alpha$-nDCG@5 metrics}
\label{tab:tab2}
\end{table}

\fontsize{9.0pt}{10.0pt}
\bibliographystyle{aaai}
\bibliography{main}

\end{document}